\documentclass[10pt,twocolumn,letterpaper]{article}

\usepackage{cvpr}
\usepackage{times}
\usepackage{epsfig}
\usepackage{graphicx}
\usepackage{amsmath}
\usepackage{amssymb}
\usepackage{caption,subcaption}
\usepackage{multirow}
\usepackage{booktabs}

\newcommand*\rot{\rotatebox{90}}

\makeatletter
\newif\if@restonecol
\makeatother

\usepackage[linesnumbered,ruled,vlined]{algorithm2e}
\usepackage{algpseudocode}
\usepackage{amsmath}

\cvprfinalcopy 

\def\cvprPaperID{5097} 

\ifcvprfinal\fi
\begin{document}

\title{Adversarial Style Mining for One-Shot Unsupervised Domain Adaptation}

\author{Yawei Luo$^{1,2}$,\hspace{2mm} Ping Liu$^{2}$,\hspace{2mm} Tao Guan$^{1,4}$,\hspace{2mm} Junqing Yu$^{1,5}$,\hspace{2mm} Yi Yang$^{2,3}$ \vspace{0.5cm} \\ 
$^1$School of Computer Science \& Technology, Huazhong University of Science \& Technology\\
$^2$ReLER, University of Technology Sydney ~ $^3$Baidu Research ~ $^4$Farsee2 Tech. Co.\\ 
$^5$Center of Network and Computation, Huazhong University of Science \& Technology 
}

\maketitle

\begin{abstract}
  We aim at the problem named One-Shot Unsupervised Domain Adaptation. Unlike traditional Unsupervised Domain Adaptation, it assumes that only one unlabeled target sample can be available when learning to adapt. This setting is realistic but more challenging, in which conventional adaptation approaches are prone to failure due to the scarce of unlabeled target data. To this end, we propose a novel Adversarial Style Mining approach, which combines the style transfer module and task-specific module into an adversarial manner. Specifically, the style transfer module iteratively searches for harder stylized images around the one-shot target sample according to the current learning state, leading the task model to explore the potential styles that are difficult to solve in the almost unseen target domain, thus boosting the adaptation performance in a data-scarce scenario. The adversarial learning framework makes the style transfer module and task-specific module benefit each other during the competition. Extensive experiments on both cross-domain classification and segmentation benchmarks verify that ASM achieves state-of-the-art adaptation performance under the challenging one-shot setting.
\end{abstract}

\section{Introduction}
Deep networks have significantly improved state of the art for a wide variety of machine-learning problems and applications. Nevertheless, these impressive gains in performance come with a price of massive amounts of manual labeled data. 
A popular trend in the current research community is to resort to simulated data, such as computer-generated scenes~\cite{richter2016gta5,ros2016synthia}, so that unlimited amount of automatic annotation is made available. However, this learning paradigm suffers from the shift in data distributions between the real and simulated domains, which poses a significant obstacle in adapting predictive models in the source domain to the target task.
The introduction of Domain Adaptation (DA) techniques aims to mitigate such performance drop when a trained agent encounters a different environment. 
By bridging the distribution gap between source and target domains, DA methods have shown their effect in many cross-domain tasks such as classification~\cite{long2018transferable}, segmentation~\cite{wu2018dcan,tsai2018OutputSpace} and detection~\cite{chen2018domain}. 

Although much progress has been made for domain adaptation, most of the previous efforts assume the availability of enough amounts of unlabeled target-domain samples. However, such an assumption can not always hold since not only data labeling but also data collection itself might be challenging, if not impossible, for the target task. For example, it could be hard to acquire rare disease information with privacy or to shoot videos under extreme weather conditions. In this data-scarce scenario with a limited amount of unlabeled data from target domains, most previous DA strategies, such as distribution alignment~\cite{hoffman2017cycada,tsai2018OutputSpace}, entropy minimization~\cite{vu2019advent}, or pseudo label generation~\cite{zou2018unsupervised}, are all prone to failure. Consequently, design a specific algorithm for this realistic but more challenging learning scenario, \emph{i.e.}, one-shot unsupervised domain adaptation (OSUDA), becomes necessary. 



\begin{figure}[t]
\begin{center}
\includegraphics[width=0.99\linewidth]{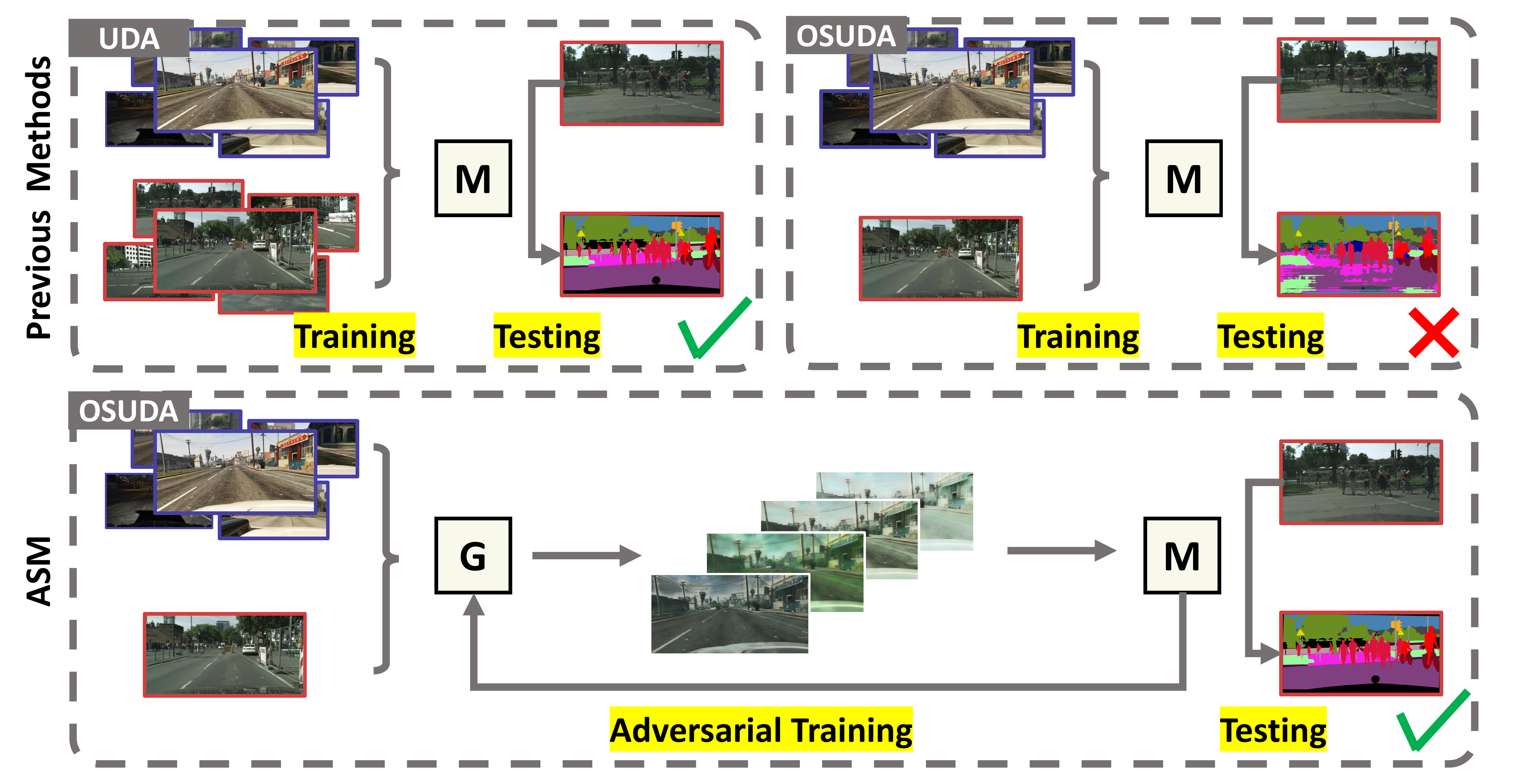}
\end{center}
\vspace{-0.3cm}
   \caption{Conventional DA methods achieve good performance in UDA task but are prone to failure under the one-shot setting. We propose ASM to deal with such challenging data-scarce scenario.}
\label{fig:brief}
\vspace{-0.5cm}
\end{figure}

Some recent works~\cite{dundar2018domain,li2019bidirectional,zhang2018fully,yue2019domain} aiming to vanilla UDA problems, try to learn style distribution in target domains based on the given unlabeled target data. The learned style distribution is utilized to translate source domain data to make them with a similar ``appearance'', \emph{e.g.}, lighting, texture, etc., as target domain data. The model trained on stylized source data can naturally, hopefully, generalize well to the target domain. However, there are a few drawbacks if directly apply those vanilla style transfer (ST) methods in One-Shot UDA settings. First, both the style transfer module and the classifier could easily over-fit due to the scarce of target domain data. With only one target sample, it is hard to learn from it to catch the actual style distribution in the target domain. Second, in those previous works~\cite{benaim2018one,cohen2019bidirectional,li2018Grad-GAN,hoffman2017cycada}, ST and DA are usually carried out in a sequential, decoupled manner, which makes it hard for ST and DA benefit mutually to each other. That means, since the ST module can not get dynamic feedback from the classifier, it might produce inappropriate stylized samples. Those inappropriate stylized samples might be either too ``hard'' or too ``easy'' for the model in the current state.

To overcome these drawbacks above, in this paper, we propose the Adversarial Style Mining (ASM) algorithm effective for OSUDA~\footnote{It should be noted that ASM can also be applied for UDA problems.}. As shown in Fig.~\ref{fig:brief}, ASM is composed of a stylized image generator $G$ and a task-specific network $M$, \emph{e.g.}, FCN for segmentation task. We design $G$ to generate arbitrary style from a sampling vector $\varepsilon$. We can change the style of the generated image by simply modifying $\varepsilon$. Unlike previous style translation works~\cite{zhu2017cycle,huang2017arbitrary}, our $\varepsilon$ is initialized by the given single target sample and will be updated according to the dynamic feedback from $M$. The updated $\varepsilon$ is suitable for the learning ability of $M$ at that time. $M$ is trained to segment or classify the stylized images from $G$ correctly. Unlike the previous efforts~\cite{yue2019domain,dundar2018domain} that train $G$ and $M$ in a decoupled manner, we construct these two key modules as an end-to-end adversarial regime. Specifically, in our ASM, $\varepsilon$ and $M$ are iteratively updated during the training. On the one hand,  $\varepsilon$ starts from the style in the solely given target sample and constantly searches for harder stylized images for the current $M$. On the other hand, $M$ is trained based on these generated stylized images and returns its feedback, so $\varepsilon$ can be adjusted appropriately. In such an adversarial paradigm, we can efficiently produce stylized images that boost the domain adaptation, thus guiding the task model $M$ to ``see'' more possible styles in the target domain beyond the solely given sample. 

Our main contributions are summarized as follows:

\noindent
(1)~We present an adversarial style mining (ASM) system to solve One-Shot Unsupervised Domain Adaptation (OSUDA) problems. ASM combines a style transfer module and a task-specific model into an adversarial manner, making them mutually benefit to each other during the learning process. ASM iteratively searches for new beneficial stylized images beyond the one-shot target sample, thus boosting the adaptation performance in data-scarce scenario.

(2)~We propose a novel style transfer module, named Random AdaIN (RAIN), as a key component for achieving ASM. It makes the style searching a differentiable operation, hence enabling an end-to-end style searching using gradient back-propagation.

(3)~We evaluate ASM on both cross-domain classification and cross-domain semantic segmentation in one-shot settings, showing that our proposed ASM consistently achieves superior performance over previous UDA and one-shot ST approaches.

\begin{figure*}[t]
\begin{center}
\includegraphics[width=0.99\linewidth]{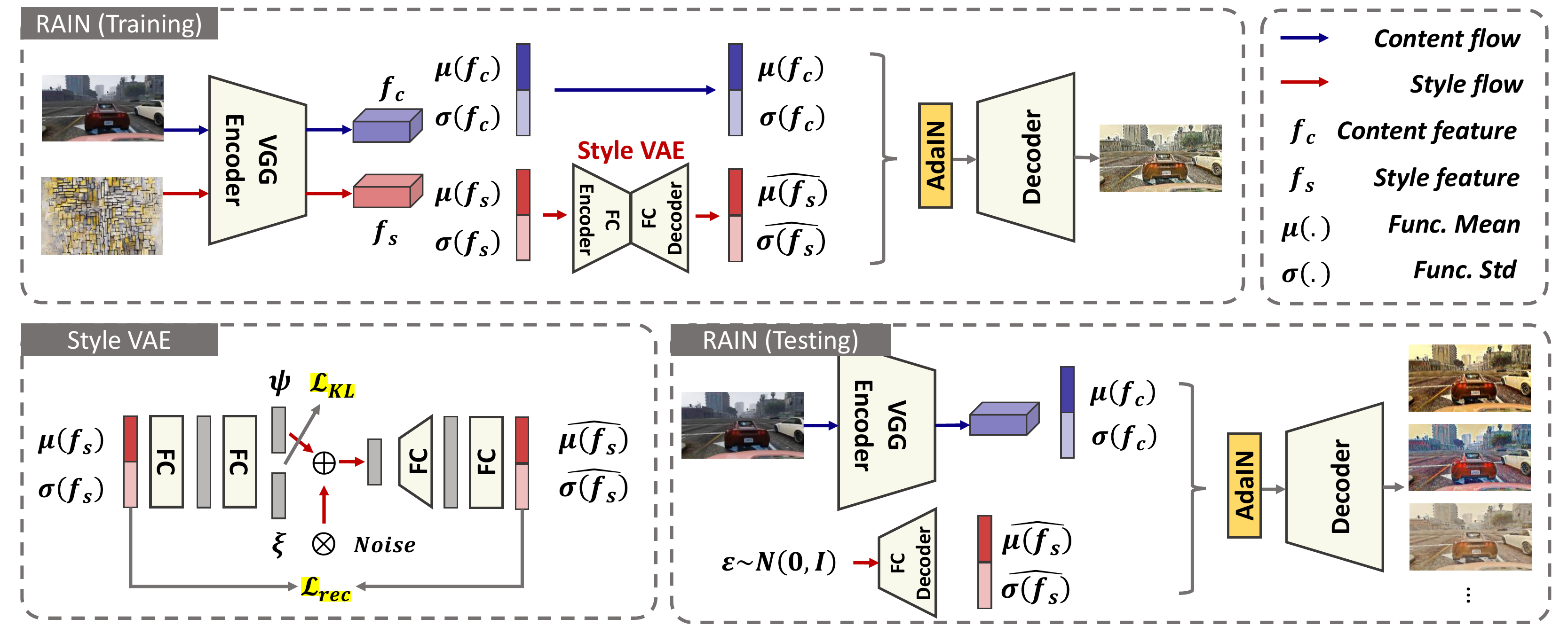}
\end{center}
   \caption{Overview of our proposed ``Random AdaIN (RAIN)'' module (See Top). Vanilla AdaIN regards each ``style'' as a pair of ``mean $\mu(f_s)$'' and ``variation $\sigma(f_s)$'' of the style image features $f_s$. Based on the vanilla AdaIN, we employ an extra VAE (See Left Bottom) in the latent space to encode the ``style'' (\emph{i.e.}, $\mu(f_s)$ and $\sigma(f_s)$) into a standard distribution. In the testing stage (See Right Bottom), on the one hand, RAIN enables us to generate arbitrary new styles from some sampled vectors $\varepsilon$, without the need for style images. On the other hand, 
   we can simply generate \textbf{other reasonable styles} near $\varepsilon$ by passing a small perturbation to $\varepsilon$. These properties of RAIN makes the style searching a differentiable operation, hence enabling an end-to-end style searching using gradient back-propagation, which is the key to achieve ASM.
   }
\label{fig:rain}
\end{figure*}





\section{Related Work}\label{sec:relatedwork}

\subsection{Domain Adaptation}
Based on the theory of Ben-David \emph{et al.}~\cite{ben2010theory}, the majority of recent DA works~\cite{long2018conditional} lay emphasis on how to minimize the domain divergence. Some methods~\cite{hoffman2016fcns,liu2016coupled,kim2017relations} aim to align the latent feature distribution between two domains, among which the most common strategy is to match the marginal distribution within the adversarial training framework~\cite{luo2019significance,tzeng2017adversarial}. More similar to our method are approaches based on the image-to-image translation aiming to make images indistinguishable across domains, where an incomplete list of prior work includes~\cite{li2018Grad-GAN,zhang2018fully,li2019bidirectional,gong2019dlow}. Joint consideration of image and feature level DA is studied in~\cite{hoffman2017cycada}. Besides alignment the latent features, Tsai \emph{et al.}~\cite{tsai2018OutputSpace} found that directly aligning the output space is more effective in semantic segmentation. Based on the output space alignment, Vu \emph{et al.}~\cite{vu2019advent} further leveraged the entropy minimization to minimize the uncertainty of predictions in target data. Another popular branch is to extract confident target labels and use them to train the model explicitly~\cite{zou2018unsupervised,zou2019confidence}.

\subsection{Style Transfer}
Style transfer aims at altering the low-level visual style within an image while preserving its high-level semantic content. Gatys \emph{et al.}~\cite{gatys2015neural} proposed the seminal idea to combine content loss and style loss based on the pre-trained neural networks on ImageNet~\cite{deng2009imagenet}. Based on this pioneering work, Huang \emph{et al.}~\cite{huang2017arbitrary} proposed the AdaIN to match the mean and variance statistics of the latent embedding of the content image and style image, then decoded the normalized feature into a stylized image. Another line of works~\cite{li2018Grad-GAN,zhu2017cycle,huang2018multimodal} is based on the generative adversarial network (GAN)~\cite{goodfellow2014gan}, which employs a discriminator to supervise the style transfer. Related to our settings, one-shot style transfer techniques have drawn more attention~\cite{benaim2018one,cohen2019bidirectional} recently. More similar to our method are to use the style transfer as a data augmentation strategy~\cite{yue2019domain,jackson2018style}.  However, these works usually regard the style transfer as a single module and do not consider the interaction to other tasks.

\section{Method}
In this section, we formally introduce One-Shot Universal Domain Adaptation (OSUDA) problem and our overall idea to address it. Then we introduce the detailed network design and the cost function to achieve the idea.

\subsection{Problem Settings and Overall Idea}\label{sec:overall idea}
We illustrate the problem setting at first. In the training process of OSUDA, we have access to the source data $X_S$ with labels $Y_S$, but only \textit{one} unlabeled target data $x_T \in X_{T}$. The goal is to learn a model $M$ based on those data to correctly predict the labels for the target domain. 

Overall, we suggest guiding the task-specific model $M$ to explore more possible styles in the target domain beyond the solely given sample. Introducing slight noise to the style from the given target sample is not an ideal solution in this case since it can not avoid overfitting. Purely randomly generating various styles, on the other hand, would produce images that are excessively deviated from the target distribution, which are either unrealistic or too hard for $M$. In our method, we propose to regard the style of the solely given target sample as an ``anchor style''.  Starting from this anchor style, images with more various styles are generated by a stylized image generator $G$, whose details will be illustrated in the subsection~\ref{subsec:RAIN}. Those generated images with new styles are utilized to boost the generalization of the task-specific model $M$. The updated $M$ provides dynamic feedback for $G$ to generate images with new styles. Comparing to the old style, those new styles are ``harder'' for the $M$ to adapt and therefore provide stronger supervision to achieve a stronger $M$.
In the following sections, we will introduce the detailed network design.


\subsection{Random AdaIN}
\label{subsec:RAIN}
We first propose a module named Random Adaptive Instance Normalization (RAIN) as the stylized image generator $G$, which can easily respond to the feedback from the task-specific model $M$. RAIN equips the AdaIN~\cite{huang2017arbitrary} with a variational auto-encoder (named style VAE) in the latent space. For the AdaIN part, similar to~\cite{huang2017arbitrary}, we employ the pre-trained VGG-19 as encoder $E$, to compute the loss function to train the decoder $D$:  
\begin{equation}
\mathcal{L}_{Adain} = \mathcal{L}_c + \lambda_s \mathcal{L}_s,
\label{eq:adain_train}
\end{equation}
where $\mathcal{L}_c$ and $\mathcal{L}_s$ denote content loss and style loss respectively, and $\lambda_s$ is a hyper-parameter controlling the relative  importance of the two losses. AdaIN re-normalizes the features of content images $f_c$ to have the same channel-wise mean and standard deviation as the features of a selected style image $f_s$ as follows: 

\begin{equation}
	\textrm{AdaIN}(f_c, f_s)= \sigma(f_s)\left(\frac{f_c-\mu(f_c)}{\sigma(f_c)}\right)+\mu(f_s),
	\label{eq:adain}
\end{equation}
where $\mu(.)$ and $\sigma(.)$ denote channel-wise mean and standard deviation operations, respectively.

The style VAE, as shown in Fig.~\ref{fig:rain}, is composed of an encoder $E_{vae}$ and a decoder $D_{vae}$, both of which contain two FC layers. $E_{vae}$ encodes $\mu(f_s) \odot \sigma(f_s)$ (where $\odot$ denotes ``concatenate'') to a Gaussian distribution $N(\psi, \xi)$\footnote{We use the notation $\psi$ for mean and $\xi$ for standard deviation in Style VAE, in order to avoid confusion to the $\mu(.)$ and $\sigma(.)$ in AdaIN.}, and $D_{vae}$ decodes a sampling $\varepsilon$ from such distribution aiming to reconstruct the original style. Therefore, besides the conventional training scheme for AdaIN, we have two extra losses for training the Style VAE. The overall training objective for RAIN is to minimize the following loss:

\begin{figure*}[t]
\begin{center}
\includegraphics[width=0.99\linewidth]{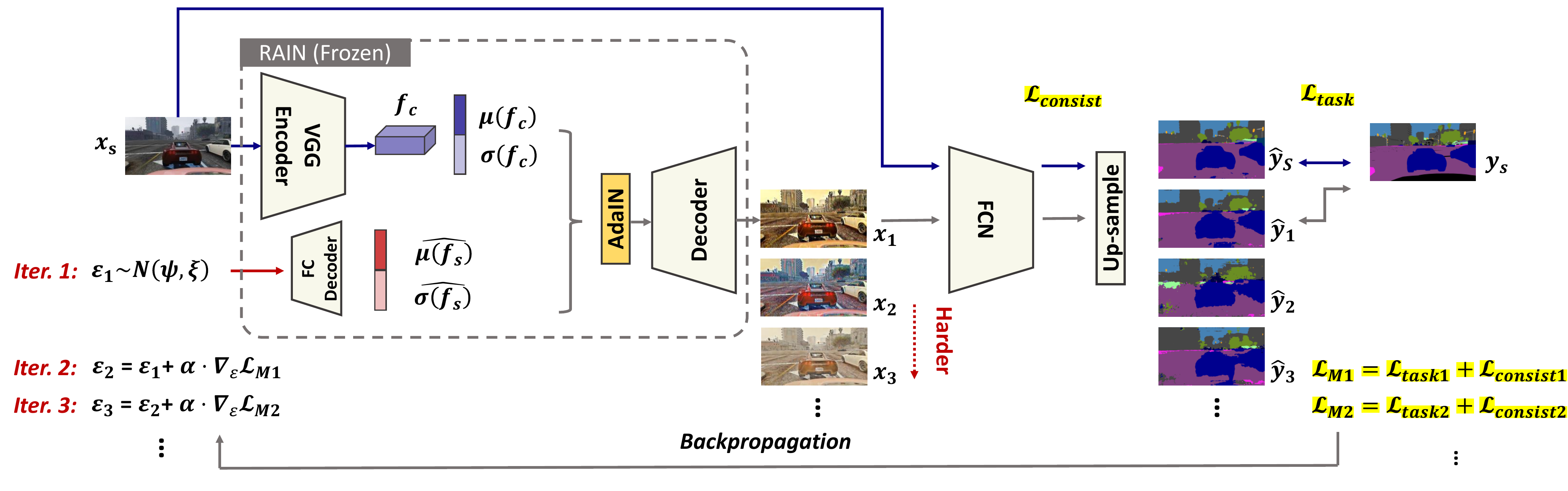}
\end{center}
\vspace{-0.5cm}
   \caption{The framework of ASM. It consists of a stylized image generator ($G$) and a task-specific network ($M$). Here we take the semantic segmentation task as an example, where $M$ can be any FCN-based structure. $G$ is a pre-trained RAIN module described in section~\ref{subsec:RAIN}. By updating its input $\varepsilon$ based on the feedback from $M$, $G$ can continually generate harder samples for $M$ to boost its generality. First, we sample a initial style vector $\varepsilon_1$ from a Gaussian distribution. In our one-shot scenario, such Gaussian distribution is defined by $\psi$ and $\xi$, which is extracted from the one-shot target image $x_t$. Second, a source domain image $x_s$, together with $\varepsilon_1$, are forwarded to RAIN to generate a stylized image $x_1$, which is then fed into $G$ to produce the training loss $\mathcal{L}_{M1}$. 
   We minimize $\mathcal{L}_{M1}$ to train $M$ to better generalize to $x_1$, and also prepare for next iteration by searching for the new vector $\varepsilon_i$ around $\varepsilon_1$ that can generate harder stylized image. Specifically, we update $\varepsilon_1$ by adding a small perturbation whose direction equals to the elements of the gradient of the loss function with respect to $\varepsilon_1$. Finally, the $\varepsilon_i$, as a harder sampling mined by ASM, will be fed into the pipeline the same way as $\varepsilon_1$ to bootstrap next iteration.}
\label{fig:mainframe}
\end{figure*}

\begin{equation}
\mathcal{L}_{RAIN} = \mathcal{L}_c + \lambda_s \mathcal{L}_s + \lambda_k \mathcal{L}_{KL} + \lambda_r \mathcal{L}_{Rec}
\label{eq:RAIN_train}
\end{equation}

Within Eq.~\ref{eq:RAIN_train}, the latter two terms form the training loss for Style VAE:

\begin{equation}
\mathcal{L}_{KL} = \textrm{KL}[\mathcal{N}(\psi, \xi)||\mathcal{N}(0, I)]
\label{eq:RAIN_kl}
\end{equation}
\begin{equation}
\mathcal{L}_{Rec} = \lVert \mu(f_s) \odot \sigma(f_s), \widehat{\mu(f_s) \odot \sigma(f_s)} \rVert_{2}
\label{eq:RAIN_rec}
\end{equation}

where $\widehat{\mu(f_s) \odot \sigma(f_s)}$ denotes the reconstructed style vector from a sampling $\varepsilon \sim \mathcal{N}(\psi, \xi)$.

\subsection{Adversarial Style Mining} \label{sec:ASM}
We illustrate ASM framework in Fig.~\ref{fig:mainframe} and the corresponding pipeline in Alg.~\ref{alg:ASM}. We employ a pre-trained RAIN module as the stylized image generator $G = \{E,D,E_{vae},D_{vae}\}$, whose parameters are kept fixed during the training. The generated images will be forwarded to the task model $M$, with the goal of leading $M$ to generalize to the target domain. Given one-shot target sample $x_T$, we can first obtain a latent distribution $\mathcal{N}(\psi, \xi)$, where $\psi, \xi$ = $E_{vae}(E(x_T))$. Each time we are given a source domain image $x_S$, we can sample $\varepsilon$ from $\mathcal{N}(\psi, \xi)$ and decode it to an initial style vector $\widehat{\mu(f_s) \odot \sigma(f_s)} = D_{vae}(\varepsilon)$, from which we can further generate an initial stylized image $x_{1}$. Since the current style of $x_{1}$ is very close to $x_T$, we can regard it as an anchor-style image. 

Following the overall idea in Sec.~\ref{sec:overall idea}, the next step is to search for some new styles around anchor style iteratively. We construct this step as an adversarial paradigm. On the one hand, we update task model $M$ by minimizing a cost function $\mathcal{L}_{M}$, training $M$ to classify (or segment) $x_{1}$ rightly. On the other hand, we update $\varepsilon$ by adding a small perturbation whose direction is consistent with the gradient of the cost function with respect to $\varepsilon$.  In this adversarial spirit, $M$ could learn to handle those more arduous samples $x_i$ around the one-shot anchor rightly, thus performing better on the unseen target domain samples. 


\begin{algorithm}[t]
  \caption{Adversarial Style Mining}
  \label{alg:ASM}
  \KwIn{Source domain data $X_S$; source domain label $Y_S$; one-shot target domain data $x_T \in X_T$; a pre-trained RAIN module $G = \{E,D,E_{vae},D_{vae}\}$; task model $M$ with parameter $\theta$; learning rate $\alpha$, $\beta$; max searching depth $n$}
  \KwOut{Optimal $\theta^*$}
  Randomly initialize $\theta$\;
  $\psi, \xi$ = $E_{vae}(E(x_T))$\;
  \For{$x_S \in X_S$}
  {
    $f_c = E(x_S)$\;
    Sampling $\varepsilon \sim \mathcal{N}(\psi, \xi)$\;
    \For{i = 1, ..., n}
    {
      Reconstruct the style vector: $\widehat{\mu(f_s) \odot \sigma(f_s)} = D_{vae}(\varepsilon)$\;
      Generate stylized image $x_{style} = D(\textrm{AdaIN}(f_c, \mu(f_c), \sigma(f_c), \widehat{\mu(f_s)}, \widehat{\sigma(f_s)}))$\;
      Update model parameters:
      $\theta \leftarrow \theta - \alpha \triangledown_{\theta}\mathcal{L}_{M}(M(x_{style}), y_S)$\;
      Update sampling:
      $\varepsilon \leftarrow \varepsilon + \beta \triangledown_{\varepsilon}\mathcal{L}_{M}(M(x_{style}), y_S)$\;
    }
  }
  return $\theta$ as $\theta^*$\;
\end{algorithm}

\subsection{Cost Function} \label{sec:Cost Function}
Two losses are used to train the task model $M$.

\textbf{Task Loss.} We employ the task loss to train $M$ to learn knowledge from source label:
\begin{equation}
\mathcal{L}_{task} = \ell (M(x_S), y_S) \; ,
\label{eq:task loss}
\end{equation}
where $x_S$ can be original or stylized source data. $\ell(\cdot, \cdot)$ is a task-specific cost function, \emph{e.g.,} multi-class cross entropy for segmentation task.

\textbf{Consistency Loss.} To further encourage task model $M$ to distill the domain invariant feature, we employ a consistency loss as follows: 
\begin{equation}
\mathcal{L}_{consist} = \frac{\sum_{i=1}^{N}\lVert z-\overline{z}\rVert_2}{N} \; ,
\label{eq:consist loss}
\end{equation}
where $z$ denotes the latent features from the second last layer of $M$, $\overline{z}$ denotes the average value of $z$ across a $N$-sized batch. The motivation behind is that a source image under different stylization should maintain similar semantic information in deep layers. Such loss constrains the semantic consistency across a mini-batch of images, which have same content but different styles.

Then the overall cost function to train $M$ is:
\begin{equation}
\mathcal{L}_{M} = \mathcal{L}_{task} + \lambda \mathcal{L}_{consist},
\label{eq:M loss}
\end{equation}
where $\lambda$ denotes a hyper-parameter controlling the relative  importance of the two losses.

\section{Experiments}
\subsection{Datasets and Evaluation Protocols}
We evaluate ASM together with several state-of-the-art UDA algorithms on both classification and segmentation tasks. We use MNIST~\cite{mnist}-USPS~\cite{usps}-SVHN~\cite{svhn} benchmarks to evaluate ASM on one-shot cross domain classification task, where MNIST (M) and USPS (U) contain images of hand-writing digits from 0 to 9 while SVHN (S) captures some images of the house number in the wild. We select three adaptation tasks, \emph{i.e.}, $M \rightarrow S$, $U \rightarrow S$ and $M \rightarrow U$, to evaluate ASM. Following the experimental setting in~\cite{tzeng2017adversarial,liu2016coupled,motiian2017few}, we use all the source domain data in the first two tasks while randomly selecting 2,000 images from MNIST in task $M \rightarrow U$. We use the classification accuracy as the evaluation metric.

For one-shot cross-domain segmentation task, we evaluate ASM on two benchmarks, \emph{i.e.}, SYNTHIA~\cite{ros2016synthia} $\rightarrow$ Cityscapes~\cite{cordts2016cityscapes} and GTA5~\cite{richter2016gta5} $\rightarrow$ Cityscapes. Cityscapes is a real-world dataset with 5,000 street scenes which are divided into a training set with 2,975 images, a validation set with 500 images and a testing set with 1,525 images. We use Cityscapes as the one-shot target domain. GTA5 contains 24,966 high-resolution images, automatically annotated into 19 classes. The dataset is rendered from a modern computer game, Grand Theft Auto V, with labels fully compatible with those of Cityscapes. SYNTHIA contains 9,400 synthetic images compatible with the Cityscapes annotated classes. We use SYNTHIA or GTA5 as the source domain in evaluation. In terms of the evaluation metrics, we leverage Insertion over Union (IoU) to measure the performance of the compared methods.


Besides the task-specific datasets, we extra leverage some data as the ``style images'' to train the RAIN. Here we follow ~\cite{huang2017arbitrary} to use a dataset of paintings mostly collected from WikiArt. However, there is no limit to choose any other website data since no annotation is required for style images.

\subsection{Implementation Details}
We use PyTorch~\cite{paszke2017automatic} for our implementation. The training process is composed of two stages. In the first stage, we use source images and style images to train the RAIN module. In the second stage, we fix RAIN and train the task model within the ASM framework. For \textbf{classification task}, we employ ResNet-18~\cite{he2016resnet} as the backbone and SGD~\cite{bottou2010SGD} as the optimizer, with a weight decay of $5e$-$4$. We train the network for a total of $30k$ iterations, with the first $600$ as the warm-up stage~\cite{Akhilesh2019Warmup} during which the learning rate increases linearly from $0$ to the initial value. Then the learning rate is divided by ten at 10k and 20k iterations. We resize the input images to $64 \times 64$ and the batch size is set to 64. For \textbf{segmentation task}, we leverage ResNet-101~\cite{he2016resnet}-based DeepLab-v2~\cite{chen2018deeplab} as the backbone of segmentor. To reduce the memory footprint, we resize the original image to $1,280 \times 720$ and random crop $960 \times 480$ as the input. We use SGD~\cite{bottou2010SGD} with a momentum of $0.9$ and a weight decay of $5e$-$4$ as the optimizer. The initial learning rates for SGD is set to $2.5e$-$4$ and is decayed by a poly policy, where the initial learning rate is multiplied by $(1 - \frac{iter}{max\_iter})^{power}$ with $power = 0.9$. We train the network for a total of $100k$ iterations, with the first $5k$ as the warm-up stage like in classification task.
In our best model, we set hyper-parameters $\lambda = 2e-4$, $\lambda_s = 1.0$, $\lambda_k = 1.0$, $\lambda_r = 5.0$, respectively. The searching depth $n$ in each iteration is set to $5$ in classification task and $2$ in segmentation task.

\subsection{Image Classification}
\begin{figure}[t]
\centering
\begin{minipage}[b]{.24\linewidth}
\centering
\includegraphics[height=1.72cm]{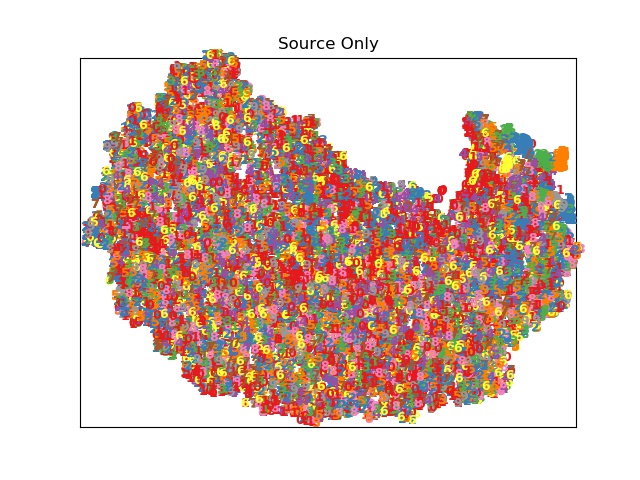}
\subcaption{}
\label{fig:tsne_Source}
\end{minipage}
\begin{minipage}[b]{.24\linewidth}
\centering
\includegraphics[height=1.72cm]{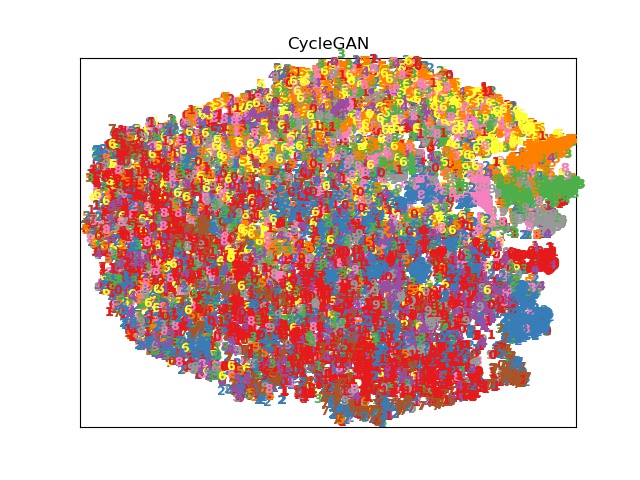}
\subcaption{}
\label{fig:tsne_CycleGAN}
\end{minipage}
\begin{minipage}[b]{.24\linewidth}
\centering
\includegraphics[height=1.72cm]{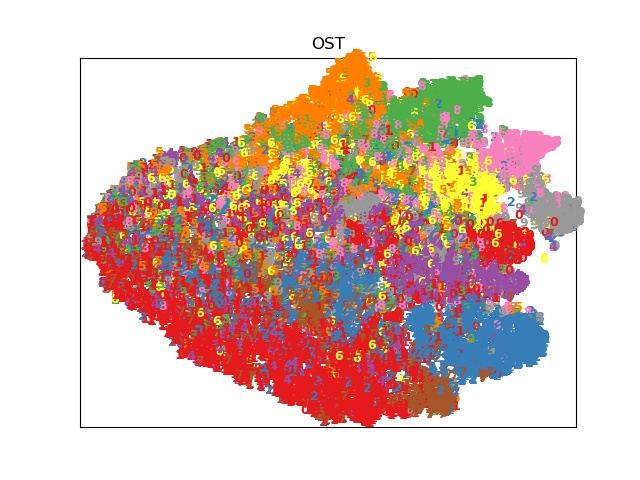}
\subcaption{}
\label{fig:tsne_OST}
\end{minipage}
\begin{minipage}[b]{.24\linewidth}
\centering
\includegraphics[height=1.72cm]{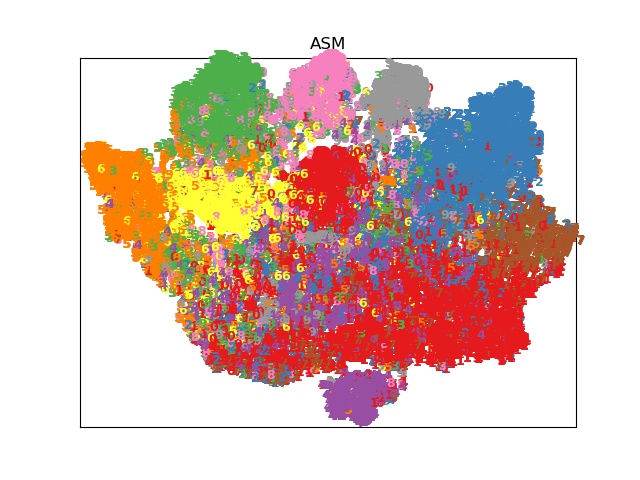}
\subcaption{}
\label{fig:tsne_ASM}
\end{minipage}
\caption{We demonstrate the effect of ASM by visualization of the learned representations in $M \rightarrow S$ task via t-SNE~\cite{maaten2008tSNE}. These results are from (a) Source only; (b) CycleGAN~\cite{zhu2017cycle}; (c) OST~\cite{benaim2018one}; (D) ASM.}
\label{fig:tsne}
\vspace{-0.5cm}
\end{figure}

In this experiment, we evaluate the adaptation scenario across MNIST-USPS-SVHN datasets. We present the adaptation results on task $M \rightarrow S$, $U \rightarrow S$ and $M \rightarrow U$ in Table~\ref{tab:MNIST-SVHN} with comparisons to the state-of-the-art domain adaptation methods. We also implement several classic style transfer methods such as CycleGAN~\cite{zhu2017cycle} and MUNIT~\cite{huang2018multimodal} under the one-shot setting. From the table, we can observe that on the task $M \rightarrow S$ and $U \rightarrow S$, ASM produces the state-of-the-art classification accuracy (46.3\% and 40.3\%), significantly outperforming other competitors under one-shot UDA settings. Moreover, ASM performs even better than the few-shot supervised methods, indicating our strategy can make the utmost of the given one-shot sample. To make our analysis more convincing, we visualize the learned representations in $M \rightarrow S$ task via t-distributed stochastic neighbor embedding (t-SNE)~\cite{maaten2008tSNE} in Fig.~\ref{fig:tsne}. Nevertheless, we can also find that all the style transfer-based methods, including CycleGAN, MUNIT, and ASM, fall short on the $M \rightarrow U$ case. Such a result is reasonable since the domain shift between $M$ and $U$ lies in the content itself but not in the style difference. This phenomenon reveals the cases that ASM and other style transfer-based methods are not applicable. 

\begin{table}[t]
\footnotesize
\caption{Cross-domain classification on MNIST-USPS-SVHN (M-U-S) datasets. $L / UL$ denotes the labeled / unlabeled data used in training. $\#TS$ denotes the number of target sample.}
\label{tab:MNIST-SVHN}
\centering
\setlength{\tabcolsep}{3.3pt}
\begin{tabular}{c|c|c|c|c|c}
\hline
$Method$    &$\#TS$ &$L/UL$  &$M \rightarrow S$  &$U \rightarrow S$ & $M \rightarrow U$\\ 
\hline
Source Only                         &-	    &-      &$20.3$ &$15.3$ &$65.4$ \\
DRCN~\cite{ghifary2016deep}         &$all$  &$UL$   &$40.1$ &-  &$91.8$ \\
GenToAdapt~\cite{san2018generate}	&$all$  &$UL$   &$36.4$	&-  &$92.5$ \\
\hline
FADA~\cite{motiian2017few}          &$10 (1/pc)$   &$L$ &$37.7$ &$27.5$ &$85.0$ \\
FADA~\cite{motiian2017few}          &$50 (5/pc)$   &$L$ &$46.1$ &$37.9$ &$92.4$ \\
\hline
CycleGAN~\cite{zhu2017cycle}        &$1$    &$UL$   &$28.2$  &$20.7$    &   $66.8$ \\
MUNIT~\cite{huang2018multimodal}    &$1$    &$UL$   &$35.0$  &$26.5$    &   $67.4$  \\
OST~\cite{benaim2018one}            &$1$    &$UL$   &$42.5$  &$34.0$    &   \bf 74.8  \\
ASM (Ours)                          &$1$    &$UL$   &\bf 46.3  &\bf 40.3    &   $68.0$  \\
\hline
\end{tabular}
\vspace{-0.3cm}
\end{table}

\begin{table*}[t]
\caption{
    Adaptation from GTA5~\cite{richter2016gta5} to Cityscapes~\cite{cordts2016cityscapes}. We present per-class IoU and mean IoU.  ``A'', ``E'' and ``P'' represent three lines of method, \emph{i.e.,} Alignment- , Entropy minimization- and Pseudo label-based DA. $\#$TS denotes the number of target sample used in training. \emph{Gain} indicates the mIoU improvement over using the source only.
    }
    \vspace{-0.5cm}
  \begin{center}
  \scriptsize
  \setlength{\tabcolsep}{3.3pt}
  \begin{tabular}{l|c|c|ccccccccccccccccccc|cc}
    \toprule
    \multicolumn{23}{c}{\textbf{GTA5 $\rightarrow$ Cityscapes}} \\
    \midrule
     &\rot{Meth.} &\rot{\#TS} & \rot{road} & \rot{side.} & \rot{buil.} & \rot{wall} & \rot{fence} & \rot{pole} & \rot{light} & \rot{sign} & \rot{vege.} & \rot{terr.} & \rot{sky} & \rot{pers.} & \rot{rider} & \rot{car} & \rot{truck} & \rot{bus} & \rot{train} & \rot{motor} & \rot{bike} & \rot{\textbf{mIoU}} & \rot{\textbf{gain}}\\ 
     \midrule
     \midrule
     
	Source only & --- & --- & 75.8 & 16.8 & 77.2 & 12.5 & 21.0 & 25.5 & 30.1 & 20.1 & 81.3 & 24.6 & 70.3 & 53.8 & 26.4 & 49.9 & 17.2 & 25.9 & 6.5 & 25.3 & 36.0 & 36.6 & ---\\
    Fully supervised &--- &---&97.9	&81.3	&90.3	&48.8	&47.4	&49.6	&57.9	&67.3	&91.9	&69.4	&94.2	&79.8	&59.8	&93.7	&56.5	&67.5	&57.5	&57.7	&68.8	&70.4 &33.8\\
    	\midrule
        \midrule
    
    CycleGAN~\cite{zhu2017cycle} &A &All &81.7 &27.0 &81.7 &30.3 &12.2 &28.2 &25.5 &27.4 &82.2 &27.0 &77.0 &55.9 &20.5 &82.8 &30.8 &38.4 &0.0 &18.8 &32.3 &41.0  &4.4 \\    
    
	AdaptSeg~\cite{tsai2018OutputSpace} & A & All &  86.5 &36.0 &79.9 &23.4 &23.3 &23.9 &35.2 &14.8 &83.4 &33.3 &75.6 &58.5 &27.6 &73.7 &32.5 &35.4 &3.9 &30.1 &28.1 &42.4& 5.8\\
    
	CLAN~\cite{Luo2019Taking} & A & All & 87.0 & 27.1 & 79.6 & 27.3 & 23.3 & 28.3 & 35.5 & 24.2 & 83.6 & 27.4 & 74.2 & 58.6 & 28.0 & 76.2 & 33.1 & 36.7 & 6.7 & 31.9 & 31.4 & 43.2 & 6.6\\ 
	
	Advent~\cite{vu2019advent} & A+E & All &89.4 &33.1 &81.0 &26.6 &26.8 &27.2 &33.5 &24.7 &83.9 &36.7 &78.8 &58.7 &30.5 &84.8 &38.5 &44.5 &1.7 &31.6 &32.4 &45.5 &8.9\\
	
	CBST~\cite{zou2018unsupervised} & P & All &86.8 &46.7 &76.9 &26.3 &24.8 &42.0 &46.0 &38.6 &80.7 &15.7 &48.0 &57.3 &27.9 &78.2 &24.5 &49.6 &17.7 &25.5 &45.1 &45.2 &8.6\\
	
	ASM (Ours) & A & All &89.8	&38.2	&77.8	&25.5	&28.6	&24.9	&31.2	&24.5	&83.1	&36.0	&82.3	&55.7	&28.0	&84.5	&45.9	&44.7	&5.3	&26.4	&31.3	&45.5   &8.9\\

	 \midrule
     \midrule
     
    CycleGAN~\cite{zhu2017cycle}    &A &One &80.3	&23.8	&76.7	&17.3	&18.2	&18.1	&21.3	&17.5	&81.5	&40.1	&74.0	&56.2	&\bf 38.3	&77.1	&30.3	&27.6	&1.7	&\bf 30.0	&22.2	&39.6   &3.0\\ 
    
	AdaptSeg~\cite{tsai2018OutputSpace} & A & One   &77.7	&19.2	&75.5	&11.7	&6.4	&16.8	&18.2	&15.4	&77.1	&34.0	&68.5	&55.3	&30.9	&74.5	&23.7	&28.3	&\bf 2.9	&14.4	&18.9	&35.2   &-1.4\\
    
	CLAN~\cite{Luo2019Taking} & A & One & 77.1	&22.7	&78.6	&17.0	&14.8	&20.5	&23.8	&12.0	&80.2	&39.5	&74.3	&56.6	&25.2	&78.1	&29.3	&31.2	&0.0	&19.4	&16.7	&37.7   &1.1\\ 
	
	Advent~\cite{vu2019advent} & A+E & One &76.1	&15.1	&76.6	&14.4	&10.8	&17.5	&19.8	&12.0	&79.2	&39.5	&71.3	&55.7	&25.2	&76.7	&28.3	&30.5	&0.0	&23.6	&14.4 &36.1&-0.5\\
	
	CBST~\cite{zou2018unsupervised} & P & One &76.1	&22.2	&73.5	&13.8	&18.8	&19.1	&20.7	&18.6	&79.5	&41.3	&74.8	&57.4	&19.9	&78.7	&21.3	&28.5	&0.0	&28.0	&13.2	&37.1   & 0.5\\
	
	OST~\cite{benaim2018one} &A &One &84.3	&27.6	&80.9	&24.1	&23.4	&26.7	&23.2	&19.4	&80.2	&\bf 42.0	&\bf 80.7	&\bf 59.2	&20.3	&84.1	&35.1	&39.6	&1.0	&29.1	&23.2	&42.3 &5.7\\

	ASM (Ours) & A & One & \bf 86.2 &\bf 35.2 &\bf 81.4 &\bf 24.2 &\bf 25.5 &\bf 31.5   &\bf 31.5 &\bf 21.9 &\bf 82.9 &30.5 &80.1 &57.3 &22.9 &\bf 85.3 &\bf 43.7 &\bf 44.9 &0.0 &26.5 &\bf 34.9 &\bf 44.5  &\bf 7.9\\
	
    \bottomrule
  \end{tabular}
  \end{center}
  \label{table:gta-cityscapes}
  \vspace{-0.5cm}
\end{table*}

\begin{table*}[t]
  \caption{
    Adaptation from SYNTHIA~\cite{ros2016synthia} to Cityscapes~\cite{cordts2016cityscapes}. We present per-class IoU and mean IoU for evaluation. ASM and state-of-the-art domain adaptation methods are compared.
    }
    \vspace{-0.3cm}
  \begin{center}
  \scriptsize
  \setlength{\tabcolsep}{6.45pt}
  \begin{tabular}{l|c|c|ccccccccccccc|cc}
    \toprule
    \multicolumn{17}{c}{\textbf{SYNTHIA $\rightarrow$ Cityscapes}} \\
    \midrule
     &\rot{Meth.} & \rot{\#TS} & \rot{road} & \rot{side.} & \rot{buil.} &  {\rot{light}} &  {\rot{sign}} & \rot{vege.} & \rot{sky} & \rot{pers.} &  {\rot{rider}} & \rot{car} &  {\rot{bus}} &  {\rot{motor}} & \rot{bike} & \rot{\textbf{mIoU}} & \rot{\textbf{gain}}
     \\ 
     \midrule
     \midrule
        
	Source only & --- & --- & 55.6 & 23.8 & 74.6 & 6.1 & 12.1 & 74.8 & 79.0 & 55.3 & 19.1 & 39.6 & 23.3 & 13.7 & 25.0 & 38.6 & ---\\
    
    Fully supervised & --- & --- & 95.1 & 72.9 & 87.3 & 46.7 & 57.2 & 87.1 &	92.1 &	74.2 & 35.0 & 92.1 & 49.3 & 53.2 &	68.8 & 70.1 & 31.5\\
    
    \midrule
	\midrule
    
	AdaptSeg~\cite{tsai2018OutputSpace} & A & All & 84.3 &42.7 &77.5 &4.7 &7.0 &77.9 &82.5 &54.3 &21.0 &72.3 &32.2 &18.9 &32.3 &46.7 & 8.1 \\
    
	CLAN~\cite{Luo2019Taking} & A & All & 81.3 & 37.0 & 80.1 & 16.1 & 13.7 & 78.2 & 81.5 & 53.4 & 21.2 & 73.0 & 32.9 & 22.6 & 30.7 & 47.8 & 9.2\\ 
	
	ADVENT~\cite{vu2019advent} & A+E & All & 85.6 &42.2 &79.7 &5.4 &8.1 &80.4 &84.1 &57.9 &23.8 &73.3 &36.4 &14.2 &33.0 &48.0   &9.4\\
	
	CBST~\cite{zou2018unsupervised} & P & All &53.6 &23.7 &75.0 &23.5 &26.3 &84.8 &74.7 &67.2 &17.5 &84.5 &28.4 &15.2 &55.8 &48.4 &9.8\\
	
	\midrule
	\midrule
	
	AdaptSeg~\cite{tsai2018OutputSpace} & A & One & 64.1	&25.6	&75.3	&4.7	&2.7	&77.0	&70.0	&52.2	&20.6	&51.3	&22.4	&19.9	&22.3	&39.1   &0.5\\
    
	CLAN~\cite{Luo2019Taking} & A & One & 68.3	&26.9	&72.2	&5.1	&5.3	&75.9	&71.4	&54.8	&18.4	&65.3	&19.2	&22.1	&20.7	&40.4   &1.8\\ 
	
	ADVENT~\cite{vu2019advent} & A+E & One & 65.7	&22.3	&69.2	&2.9	&3.3	&76.9	&69.2	&55.4	&21.4	&77.3	&17.4	&21.4	&16.7	&39.9   &1.3\\
	
	CBST~\cite{zou2018unsupervised} & P & One &59.6	&24.1	&72.9	&5.5	&13.8	&72.2	&69.8	&55.3	&21.1	&57.1	&17.4	&13.8	&18.5	&38.5   &-0.1\\
	
	OST~\cite{benaim2018one} & A & One &\bf75.3	&\bf31.6	&72.1	&\bf12.3	&9.3	&76.1	&71.1	&51.1	&17.7	&68.9	&19.0	&\bf 26.3	&\bf 25.4	&42.8 & 4.7\\
	
	ASM (Ours) & A & One &73.5	&29.0	&\bf 75.2	&10.9	&\bf 10.1	&\bf 78.1	&\bf 73.2	&\bf 56.0	&\bf 23.7	&\bf 76.9	&\bf 23.3	&24.7	&18.2	&\bf 44.1   &\bf 6.0\\

    \bottomrule
  \end{tabular}
  \end{center}
  \label{table:synthia-cityscapes}
  \vspace{-0.8cm}
\end{table*}

\subsection{Semantic Segmentation}
For the cross-domain segmentation task, we compare our method with several recent UDA methods, including \textbf{CBST}~\cite{zou2018unsupervised}, \textbf{AdaptSeg}~\cite{tsai2018OutputSpace}, \textbf{CLAN}~\cite{Luo2019Taking}, \textbf{ADVENT}~\cite{vu2019advent}. Divided by the different strategies, these methods can be categorized into three groups: (i) alignment-based method, \emph{i.e.}, AdaptSeg, CLAN, whose idea is to make the distribution of two domains to be similar; (ii) Entropy minimization-based method, \emph{i.e.}, ADVENT, which tends to minimize the uncertainty of predictions in target data; and (iii) Pseudo label-based method, \emph{i.e.}, CBST, which extracts confident target labels and use them to train the model explicitly. For a clear comparison, we also report the segmentation result when using the source data only or using all the labeled target data to train the model. As we can observe, there is a large performance gap ($36.6\%$ vs $70.4\%$) between the two approaches. 

Firstly, we evaluate these methods under the conventional UDA settings that all the unlabeled target data are available. As shown in Table~\ref{table:gta-cityscapes}, these conventional UDA strategies can give a huge boost to the source-only baseline, bringing at least $5\%$ improvement in terms of mIoU. Although we mainly concern about OSUDA in this paper, we also evaluate ASM under conventional UDA settings. In such a data-rich scenario, we can extract variant anchor styles $\{\psi_i, \xi_i\}$ to help ASM to exploit more possible styles in the target domain. Accordingly, ASM yields $45.5\%$ in terms of mean IOU, which is on par with other methods and slightly better than it performs under the one-shot setting.

Secondly, we compare ASM with the above methods under the One-shot UDA setting. Not surprisingly, all the competitors deteriorate significantly in such a data-scarce scenario. Some of them even yield worse mIoU than the source only baseline due to the overfitting to the One-shot target sample. Besides the UDA methods, we also compare our method with state-of-the-art one-shot style transfer methods, \emph{e.g.,} OST~\cite{benaim2018one} and CycleGAN~\cite{zhu2017cycle}. To fairly compare these methods with ASM, we additionally train a ResNet-101-based segmentor upon the generated samples from OST or CycleGAN. We find that both CycleGAN and OST can improve the mIoU over the source only baseline, proofing that the style transfer is a robust strategy facing the data-scarce scenario. Furthermore, ASM boosts the mIoU to a new benchmark of $44.5\%$, which demonstrates the advantage of our adversarial scheme in ASM over the sequential combination of style transfer and segmentation like OST and CycleGAN.


Finally, by comparing the performance of ASM under UDA and OSUDA settings, we can observe that reducing the visible target data does not hurt ASM ($45.5\% \rightarrow 44.5$\%) as much as it hurts the other competing methods ($\sim 44.0\% \rightarrow \sim 37.0$\%). The smaller performance drop between One-Shot and conventional settings further proves that ASM can efficiently search for useful styles from the solely given samples. Such a self-mining mechanism minimizes the impact of missing target data. The same observation can be also found in SYNTHIA $\rightarrow$ Cityscapes task (See Table~\ref{table:synthia-cityscapes}). 

\begin{figure*}[htb]
\centering
\includegraphics[width=0.99\linewidth]{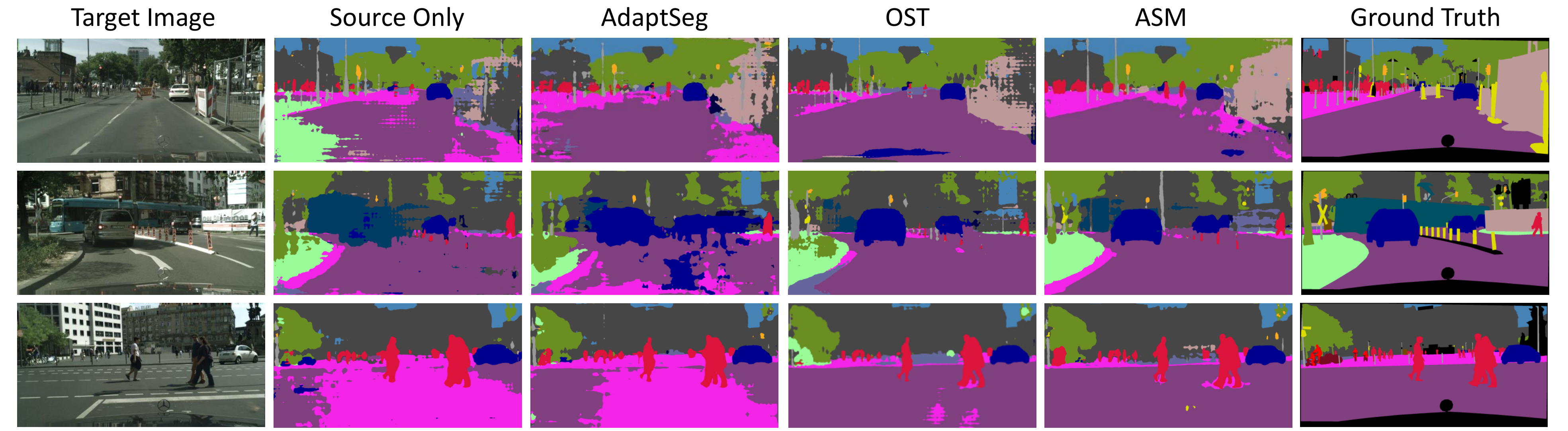}
\caption{Qualitative results of One-Shot UDA segmentation for GTA5 $\rightarrow$ Cityscapes. For each target image, we show the non-adapted (source only) result, adapted result with AdaptSeg~\cite{tsai2018OutputSpace}, OST~\cite{benaim2018one}, ASM (ours) and the ground truth label map.}
\label{fig:result}
\end{figure*}

\begin{figure}[htb]
\begin{center}
\includegraphics[width=0.99\linewidth]{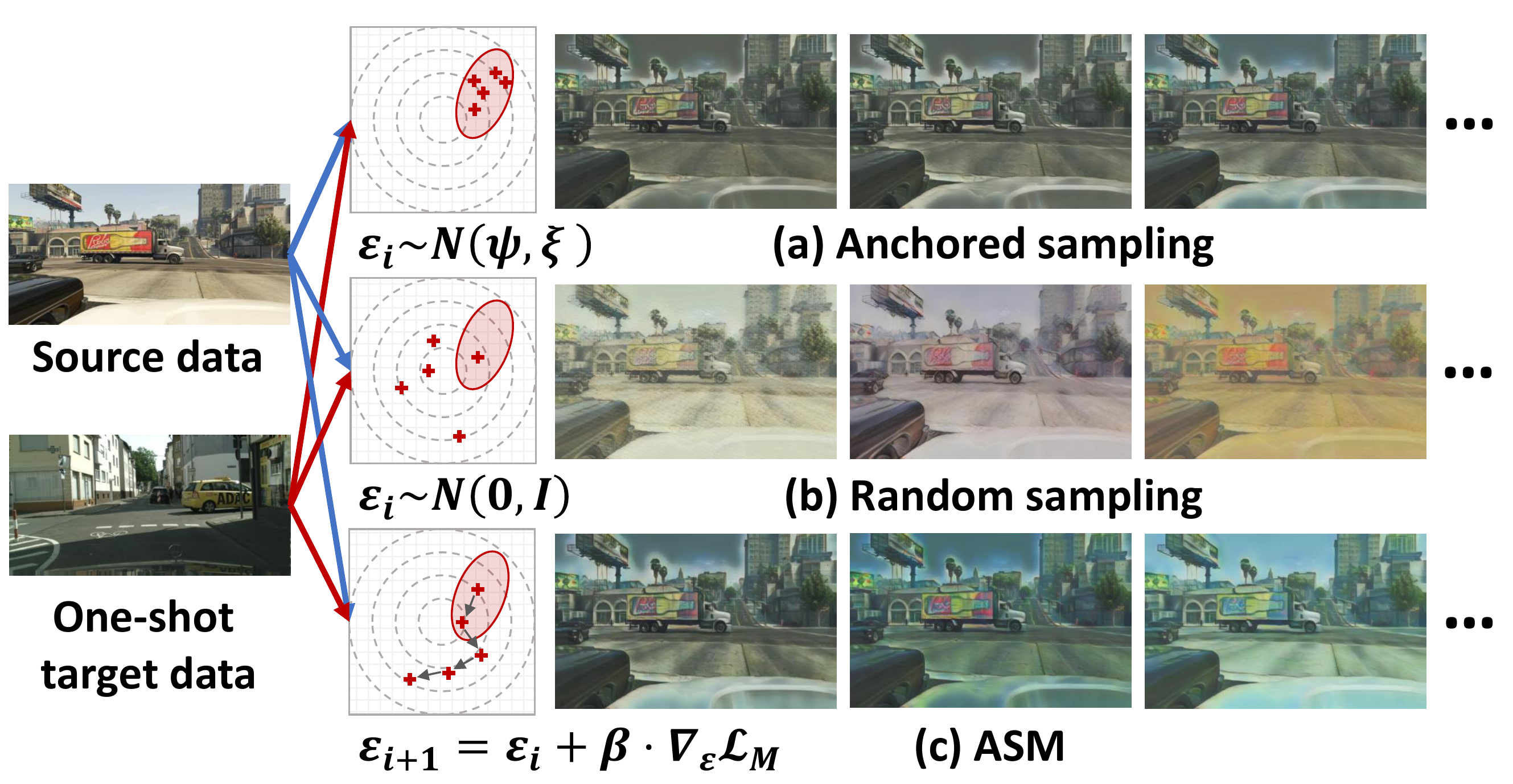}
\end{center}
   \caption{The comparison of different sampling strategies, where we visualize their respective generated images.}
\label{fig:sampling}
\vspace{-0.5cm}
\end{figure}

\subsection{Analysis of the proposed method}
\begin{figure}[t]
\centering
\includegraphics[height=3cm]{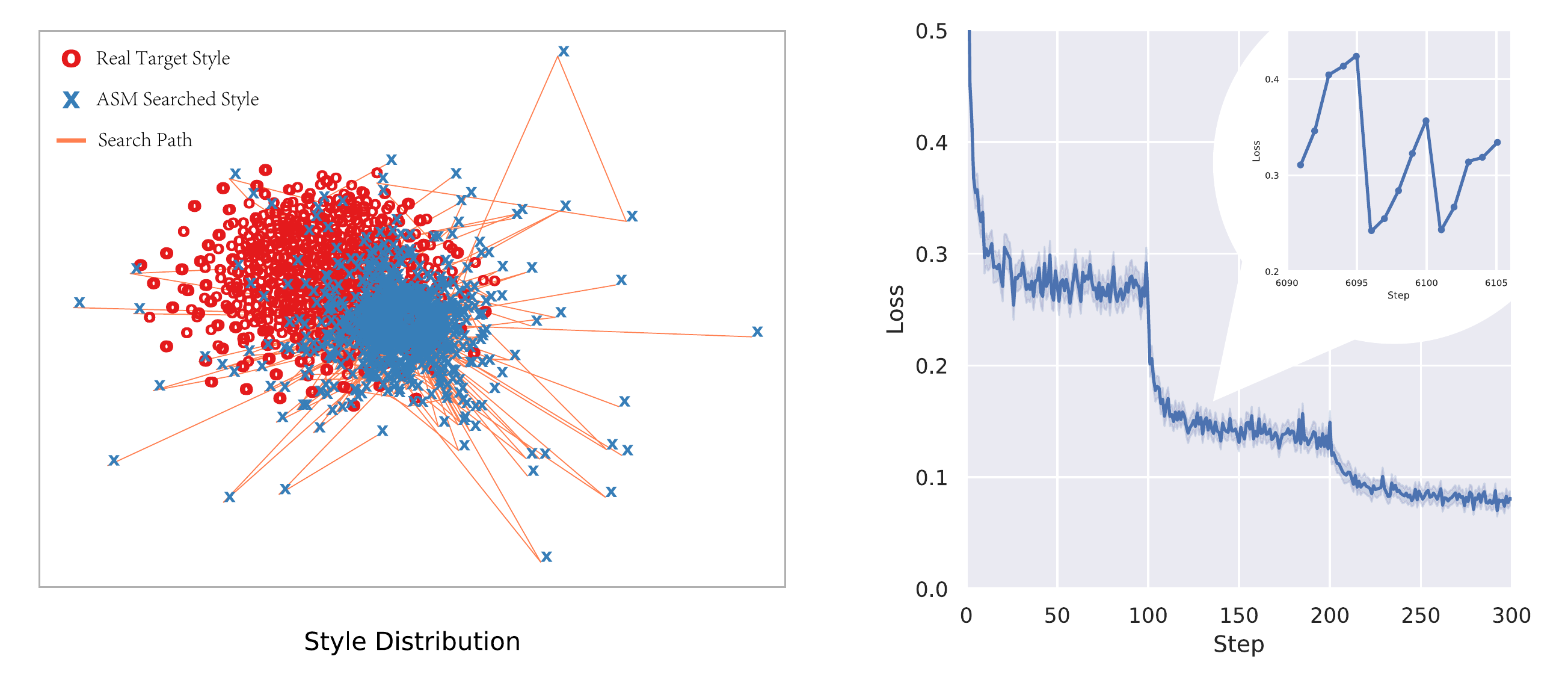}
\caption{\textbf{Left}: The style distribution of real target domain (Red) and searched by ASM (Blue). Yellow lines represent the search paths. \textbf{Right}: The training loss of ASM.}
\label{fig:analysis}
\end{figure}

\textbf{Training Stability.} Apart from the conventional adversarial training paradigm between two sub-networks, the adversaries in ASM are a network $M$ and a sampling vector $\varepsilon$. This experiment aims to explore whether these two can constitute an effective and stable adversarial system. We employ the training loss $\mathcal{L}_M$ as a proxy to evaluate the training stability of ASM (See Fig.~\ref{fig:analysis} Right.). From a local perspective, $\mathcal{L}_M$ keeps increasing within each mining iteration (the mining depth is 5 in this experiment), indicating that the sampling strategy of $\varepsilon$ can iteratively produce harder stylized images for $M$. From a global perspective, $\mathcal{L}_M$ converges to a small value at the end of training, indicating that $M$ ultimately learns to generalize to these more arduous styles around the given anchor style. Combining the local and global aspects, we can conclude that ASM is an effective and stable adversarial course.

\textbf{Style Distribution.} Here we analyze the distribution of new styles explored by ASM. Obviously, we hope that the new samples searched by ASM can overlap the real style distribution in the target domain. However, it is nearly impossible since only one target sample can be seen during training. Here we visualize the embedded styles in $M \rightarrow S$ task via t-SNE~\cite{maaten2008tSNE} (See Fig.~\ref{fig:analysis} Left.), where the red points denote the real target styles while the blue ones represent the styles mined by ASM. We can observe that ASM can efficiently search for ``unseen'' styles around the anchor style, thus promoting domain alignment in terms of style. 

\textbf{Sampling Strategy.} In this section we conduct the variation study on the $\varepsilon$ sampling methods. Based on the proposed RAIN module, we consider three different sampling strategies: (a) anchored sampling, \emph{i.e.}, $\varepsilon_i \sim \mathcal{N}(\psi, \xi)$; (b) random sampling, \emph{i.e.}, $\varepsilon_i \sim \mathcal{N}(0, I)$; and (c) ASM. The visualization comparison of the three sampling variants is depicted in Fig.~\ref{fig:sampling}. We also report the mIoU using these three strategies on task GTA5 $\rightarrow$ Cityscapes in Table~\ref{tab:sampling}. As shown first row in Fig.~\ref{fig:sampling}, anchored sampling would lead to very similar images near the given target sample. On the other hand, random sampling would produce many styles that are not helpful for the adaptation (See second row). Finally, the last row shows the stylized images found by ASM. From left to right, the generated style is increasingly different from the anchor style and harder for $M$. Together with the fact that ASM outperforms the former two sampling strategies by around $2\%$ in terms of mIoU, we can conclude that ASM offers better sampling strategy for the one-shot adaptation scenario.

\begin{table}[t]
\footnotesize
\caption{Segmentation performance on task GTA5 $\rightarrow$ Cityscapes, using variant of sampling strategy of $\varepsilon$.}
\label{tab:sampling}
\centering
\setlength{\tabcolsep}{5.3pt}
\begin{tabular}{c|c|c|c}
\toprule
\textbf{Sampling}  &Anchored  & Random    & ASM\\ 
\midrule
\textbf{mIoU}      &42.9       & 42.4      & \textbf{44.5}\\
\bottomrule
\end{tabular}
\vspace{-0.5cm}
\end{table}

\section{Conclusion} \label{sec:conclude}
In this paper, we introduce the Adversarial Style Mining (ASM) approach, aiming at the unsupervised domain adaptation (UDA) problem in case of a target-data-scarce scenario. ASM combines the style transfer module and the task model in an adversarial manner, iteratively and efficiently searching for new stylized samples to help the task model to adapt to the almost unseen target domain. ASM is general in the sense that the task-specific sub-network $M$ can be changed according to different cross-domain tasks. Experimental results on both classification and segmentation tasks validate the effectiveness of ASM, which yields state-of-the-art performance compared with other domain adaptation approaches in the one-shot scenario.

{\small
\bibliographystyle{ieee_fullname}
\bibliography{egpaper_for_review}
}

\end{document}